\documentclass{article}

\usepackage{spconf,amsmath,graphicx}
\usepackage{amsfonts,amssymb}
\usepackage{booktabs}
\usepackage{subfig}
\usepackage{color}
\usepackage{hyperref}

\title{Multiclass-SGCN: Sparse Graph-based Trajectory Prediction with Agent Class Embedding}
\name{Ruochen Li, Stamos Katsigiannis, Hubert P. H. Shum\footnotemark[1]$^{*}$}

\address{Durham University, Department of Computer Science, Durham, DH1 3LE, UK}
\begin{document}

%
\maketitle
\begin{abstract}
Trajectory prediction of road users in real-world scenarios is challenging because their movement patterns are stochastic and complex. Previous pedestrian-oriented works have been successful in modelling the complex interactions among pedestrians, but fail in predicting trajectories when other types of road users are involved (e.g., cars, cyclists, etc.), because they ignore user types. Although a few recent works construct densely connected graphs with user label information, they suffer from superfluous spatial interactions and temporal dependencies. To address these issues, we propose Multiclass-SGCN, a sparse graph convolution network based approach for multi-class trajectory prediction that takes into consideration velocity and agent label information and uses a novel interaction mask to adaptively decide the spatial and temporal connections of agents based on their interaction scores. The proposed approach significantly outperformed state-of-the-art approaches on the Stanford Drone Dataset, providing more realistic and plausible trajectory predictions.

\end{abstract}
%


\section{Introduction}
\label{sec:intro}

\renewcommand{\thefootnote}{\fnsymbol{footnote}}
\footnotetext[1]{\{ruochen.li, stamos.katsigiannis, hubert.shum\}@durham.ac.uk
\\\indent Corresponding author: Hubert P. H. Shum}



Trajectory prediction has drawn considerable attention with the development of autonomous vehicles in recent years. Specifically, models take the observed trajectories of different agents in real-world scenes to predict their future movement patterns, benefiting self-driving cars for collision avoidance \cite{liu2021trajectorysurvey}, as well as anomalous movement flow detection \cite{zhou2015dynamic}. To tackle the challenge of modelling the complex and stochastic nature of social interaction patterns,
methods focusing on spatial interaction modelling and temporal dependency capturing are proposed.
Social-LSTM \cite{Alexandre2016lstm} uses pooling windows for interaction modelling and recurrent architecture for temporal capturing, whereas Social-STGCNN \cite{Mohamed2020socialstgcnn} uses relative distance to measure interactions between agents and temporal convolution networks (TCN) \cite{bai2018tcn} to handle temporal dependencies. STAR \cite{YuMa2020Spatio} and TF \cite{giuliari2020Transformer} propose transformer-based \cite{Vaswani2017Transformer} architectures for both spatial and temporal aspects, achieving impressive performance. As densely connected graphs may generate superfluous interactions, leading to impractical computational costs, Sparse Graph Convolution Network (SGCN) \cite{shi2021sgcn} proposes a self-attention based sparse graph architecture to mitigate these problems.

\begin{figure*}
  \includegraphics[width=\textwidth,height=4cm]{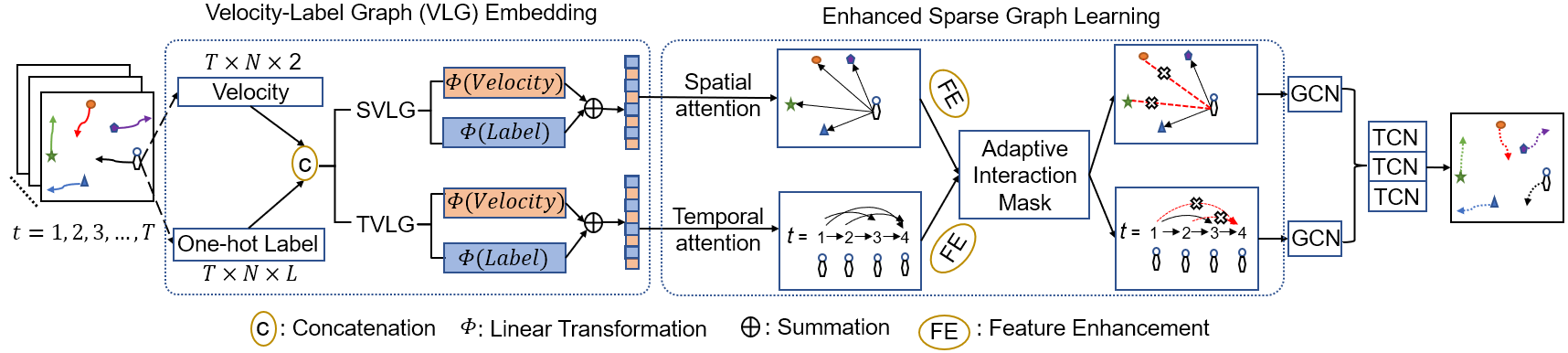}
  \caption{The network structure of Multiclass-SGCN. Given a sequence of $T$ frames including $N$ agents, we extract the velocity and label features to build spatial and temporal velocity-label graph (SVLG and TVLG). The embedded VLG features are passed into enhanced sparse graph learning with the proposed adaptive interaction mask to construct meaningful sparse attention adjacency matrices. 
  Graph convolution networks (GCN) and TCN are employed to aggregate and make predictions.}
  \label{fig:overview}
\end{figure*}


The main challenge of trajectory prediction is to consider the different movement behaviours of different classes of agents. The aforementioned research only focuses on pedestrians and does not consider other classes of agents, such as cars and cyclists, which have a significant effect on trajectory prediction. Intuitively speaking, even if two agents have a similar velocity, human instincts would force us to pay more attention to the movements of the larger agents, such as considering car over bicycle. To address this issue, Semantics-STGCNN \cite{rainbow2021semanticsstgcnn,men21pytorch} considered class labels for multi-class trajectory prediction by embedding agent-label features into the velocity representations \cite{zhang2019semanticskeleton}, ensuring that the upcoming GCN \cite{kipf2016gcn} aggregates both features. Nevertheless, Semantics-STGCNN still suffers from the superfluous interactions problem as it uses a densely connected graph. It also lacks a separate modelling of temporal dependencies, thus suffering for long-term predictions.


In this paper, we propose Multiclass Sparse Graph Convolution Network (\textit{Multiclass-SGCN}), an attention-based sparse GCN for multi-class trajectory prediction  that models interactions and temporal dependencies among multi-class agents in real scenes. 
We introduce a novel method to embed the correlated agent label and velocity features to build the velocity-label graph (VLG) representation, with particular care to learn the optimal embedding for each feature separately. 
In the sparse graph learning module, we designed a novel adaptive interaction mask to spatially and temporally evaluate attention patterns and generate plausible sparse adjacency matrices, enabling each agent to focus only on explicit neighbours and important time steps. 
Finally,  GCN \cite{kipf2016gcn} and TCN \cite{bai2018tcn} layers are employed for the final trajectory prediction.

Performance was evaluated on the Stanford Drone Dataset (SDD) \cite{Robicquet2016SDD} against state-of-the-art approaches, showing that our proposed model outperforms all existing methods for all the examined evaluation metrics by a significant margin. We open our source code for research and validation: {\color{blue} \url{https://github.com/Carrotsniper/Multiclass-SGCN}}.

The contributions of this work are: \textbf{(1)} We present \textit{Multiclass-SGCN}, a GCN for predicting multi-class agent trajectories, which outperforms state-of-the-art methods. \textbf{(2)} To effectively model the different patterns of multi-class agent trajectories, we propose a novel algorithm to separately embed the correlated features of class label and velocity, resulting in an optimal embedding for different natures of input features. \textbf{(3)} To create sparse attention of neighbours from different classes, we propose an adaptive interaction mask that adaptively filters neighbours of lower influence.










\section{Multiclass-SGCN}
\label{sec:methodology}

Given a series of ${T}$ video frames with $N$ agents, the corresponding 2-D trajectory coordinates $(x_{t}^{i}, y_{t}^{i})$, velocity $\mathcal{V}_{t}^{i} = (x_{t}^{i} - x_{t-1}^{i}, y_{t}^{i} - y_{t-1}^{i})$, and one-hot encoded semantic labels $\mathcal{L}_{t}^{i}$, $\forall$ $t\in [1,T]$ and $\forall$ $i\in[1,N]$, the goal of multi-class trajectory prediction is to predict the future trajectory coordinates of each agent $(x_{t}^{i}, y_{t}^{i})$ $\forall$ $t\in [T+1,T']$. An overview of the proposed Multiclass-SGCN for trajectory prediction is provided in \figurename~\ref{fig:overview}. We employ SGCN \cite{shi2021sgcn} as our backbone as it introduces a self-attention mechanism to enhance the spatial and temporal sparsity of the neighbour graph. The two key components of our network are the velocity label graph embedding that separately embeds the velocity and class labels for an optimal representation, and the enhanced sparse graph learning that adaptively determines the neighbour graph for each agent based on its attention preferences.

\subsection{Velocity-Label Graph (VLG) Embedding}
\label{sec:embedding}
We observe that the two important factors that affect the movement of an agent are the classes and velocity of neighbours. Class labels, $\mathcal{L}_t^i$, can indicate how different classes of agents, such as pedestrian, car, cyclist, have different influences \cite{rainbow2021semanticsstgcnn}. Velocity, $\mathcal{V}_{t}^{i}$, enhances the ability of a model to capture the geometric features of agents \cite{Mohamed2020socialstgcnn}. As velocity and classes are highly correlated, such as a car would have a higher speed, it would be advantageous to model them together. At the same time, as they are two different features, it would be better to embed them separately.


To encode the spatial and temporal features, we construct a spatial VLG (SVLG) and a temporal VLG (TVLG). SVLG contains the features of all the agents at time step $t$, with $G_{svlg} = (X_t, A_{t})$, $X_{t} = \{ \mathbf{x}_{t}^{i} \mid i = 1,...,N\}$, while TVLG contains the features of each individual agent over all time steps, such that $G_{tvlg} = (X^{i}, A^{i})$, $X^{i} = \{ \mathbf{x}_{t}^{i} \mid t = 1,...,T\}$. $X$ is the concatenation of $\mathcal{V}_{t}^{i}$ and $\mathcal{L}_{t}^{i}$, and $A_{t}$ and $A^{i}$ are adjacency matrices that represent the edges of the SVLG and TVLG respectively, indicating whether the nodes are connected (denoted as 1) or not (denoted as 0). Following \cite{shi2021sgcn}, $A^i$ is initialised as 1 and $A_t$ as an upper triangular matrix filled with 1.

We propose a \emph{velocity-label graph (VLG) embedding} that combines the advantages of velocity and class label, while learning an optimal embedding for each of them.
The graph embedding of VLG is computed by combining the embeddings of velocities and one-hot encoded class labels of agents:
\begin{equation} 
    \begin{split}
    E_{vlg} = E_{vlg}^{\mathcal{V}} + E_{vlg}^{\mathcal{L}} \\
    E_{vlg}^{\mathcal{V}} = \phi (G_{vlg}^{\mathcal{V}} , W_{E_{vlg}^{\mathcal{V}}}) \\
    E_{vlg}^{\mathcal{L}} = \phi (G_{vlg}^{\mathcal{L}} , W_{E_{vlg}^{\mathcal{L}}})
    \end{split}
    \label{eq:vlg}
\end{equation}
where $G_{vlg}^{\mathcal{V}}$ and $G_{vlg}^{\mathcal{L}}$ are subgraphs of VLG corresponding to the velocity and label features respectively,
$\phi(\cdot , \cdot)$ a linear transformation, $W_{E_{vlg}^{\mathcal{V}}} \in \mathbb{R}^{2 \times D_{E_{vlg}}}$ and $W_{E_{vlg}^{\mathcal{L}}} \in \mathbb{R}^{L \times D_{E_{vlg}}}$ the weights of the linear transformation, $L$ the length of encoded one-hot labels, and $D_{E_{vlg}}$ the embedding size.

\subsection{Enhanced Sparse Graph Learning}
\label{sec:aim}
We enhance the sparse graph learning module of SGCN \cite{shi2021sgcn} to better model the multi-class nature of the problem. This module is constructed from the numerical interaction scores calculated by the self-attention module. It then extracts high-level spatial-temporal interaction features and uses an interaction mask with a fixed threshold of $0.5$ to optimise the sparsity of graph representations by removing inexplicit connections. We argue that the interaction mask threshold should be adaptively adjusted through the learning process of each individual agent. 

 
 
 

Given the embedded SVLG and TVLG, $E_{svlg}$ and $E_{tvlg}$,
a self-attention module  \cite{Vaswani2017Transformer} is implemented to calculate the attention scores $\mathcal{A}$ between each node pairs:
\begin{equation}
    \begin{split}
        Q_{vlg} = \phi (E_{vlg}, W_{Q}^{vlg}), \hspace{0.5cm}
        K_{vlg} = \phi (E_{vlg}, W_{K}^{vlg}) \\
        \label{eq:attention}
        \mathcal{A}_{vlg} = \text{Softmax}(\frac{Q_{vlg}\times K_{vlg}^{T}}{\sqrt{d_{vlg}} }) \\
    \end{split}
\end{equation}
where $\phi (\cdot,\cdot)$ denotes a linear transformation, $W_{Q}^{vlg}$ and $W_{K}^{vlg}$ are learnable weight matrices, $\sqrt{d_{vlg}}$ is the scaled factor for numerical stability. The output spatial and temporal attention matrices,  $\mathcal{A}_{svlg}$ and $\mathcal{A}_{tvlg}$, are of size $T\times N\times N$ and $N\times T\times T$, respectively. 
Following \cite{shi2021sgcn}, we implement a feature enhancement module using a series of asymmetric convolution layers \cite{Szegedy2015AsyConv} to extract high-level interaction features, and using one-by-one convolutions on the spatial attention scores to capture the temporal dependencies, thus creating the high-level interaction attention features $F_{svlg}$ and $F_{tvlg}$. 


To sparsify the high-level interaction attention matrix, we propose an \emph{adaptive interaction mask (AIM)} to extract the set of neighbours in SVLG and TVLG.
Manually-set fixed interaction thresholds, as used by SGCN \cite{shi2021sgcn}, cannot fully describe the patterns of spatial interactions and temporal dependencies of each agent. 
We propose an average operator to adaptively calculate a threshold and remove the influence of less important neighbours, allowing the system to adapt according to the interactions of various types of agents, thus being more suitable for more complex scenes compared to the global threshold approach of SGCN \cite{shi2021sgcn}.
In particular, the $(i,j)$-th element of the adaptive sparse interaction mask $M_{vlg}$ is computed as: 
\begin{equation}
    M_{vlg}[i,j] =
    \begin{cases}
    1, \; \sigma (F_{vlg}[i,j]) > \frac{\sum_{j=1}^N \sigma (F_{vlg}[i,j])}{N} \\
    0, \; \text{otherwise}
    \end{cases}
\end{equation} 
where $\sigma$ indicates the Sigmoid function. Using the adaptive interaction mask, we construct a sparse adjacency matrix for graph convolution, and because of the removal of superfluous connections, the sparse graph enables the GCN model to learn from influential neighbours, thus improving both training speed and prediction accuracy.
Similarly to \cite{shi2021sgcn}, we apply two separate branches of the GCN \cite{kipf2016gcn} to fuse the sparse spatial VLG and sparse temporal VLG. The two GCN branches differ in the order of their input, as the first is fed the spatial VLG before the temporal VLG, whereas the second is fed in the reverse order. Then, the last outputs of these two GCN branches are summed to provide the final trajectory representation $H$.
Finally, temporal convolution networks (TCN) \cite{bai2018tcn} are used on the temporal dimension, assuming that the coordinates $(x_{t}^{i}, y_{t}^{i})$ of agent $i$ at frame $t$ follow the bi-variate Gaussian distribution as $N(\mu_{t}^{i}, \sigma_{t}^{i}, \rho_{t}^{i})$, a cascade of TCN layers can be used to predict parameters in the bi-variate Gaussian distribution. To train the proposed network, we minimise the negative log-likelihood loss function to estimate the trained parameters follow \cite{Mohamed2020socialstgcnn}.

\section{Experimental Results}
\label{sec:results}

The proposed model was trained and validated on the Stanford Drone Dataset (SDD) \cite{Robicquet2016SDD}.
SDD has class labels for six different types of agents, including \textit{pedestrian}, \textit{cyclist}, \textit{cart}, \textit{car}, \textit{skater}, and \textit{bus}. Data is captured from bird's-eye view by flying a drone over Stanford University's campus. We existing works \cite{shi2021sgcn}, \cite{gupta2018social} that apply 8 observed frames (3.2 seconds) to predict the next 12 frames (4.8 seconds), then 20 samples are derived from the learnt multivariate distribution. The model was evaluated in terms of the Minimum Average Displacement Error (mADE) and the Minimum Final Displacement Error (mFDE) as in \cite{Mohamed2020socialstgcnn}, as well as in terms of the Average ADE (aADE) and the Average FDE (aFDE) proposed by \cite{rainbow2021semanticsstgcnn} who argued that aADE and aFDE evaluate the models more holistically. The Adam \cite{kingma2014method} optimiser was used for training, with a 0.0001 learning rate and a batch size of 256. 
To compare with Semantics-STGCNN \cite{rainbow2021semanticsstgcnn}, we also normalised and denormalised the input trajectory data with a scaling factor of 10. Training typically converged in around 35-45 epochs.

\subsection{Quantitative Results}

The proposed method was compared to 8 models in total, including the baseline Linear model, energy function based behavioral model (SF \cite{Yamaguchi2011bm} ), Social-LSTM \cite{Alexandre2016lstm}, Social-GAN~\cite{gupta2018social}, CAR-Net \cite{sadeghian2017carnet}, DESIRE \cite{lee2017desire}, Social-STGCNN \cite{Mohamed2020socialstgcnn} and Semantics-STGCNN \cite{rainbow2021semanticsstgcnn}, the existing state-of-the-art model for multi-class trajectory prediction. Notably, the results of Semantics-STGCNN were evaluated using the published source code, whereas other results were provided by~\cite{rainbow2021semanticsstgcnn}. Results are presented in \tablename~\ref{tab:allmodels} in terms of mADE and mFDE. It is evident that the proposed model outperformed all other models, including the latest Semantics-STGCNN \cite{rainbow2021semanticsstgcnn} with a 3.76 decrease in mADE and 3.71 decrease in mFDE, indicating the importance of considering label information and velocity in complex trajectory prediction tasks, as well as of using an adaptive interaction mask. Furthermore, as discussed in \cite{rainbow2021semanticsstgcnn}, common minimum-based metrics (mADE and mFDE) focus only on the best sampled sample, which is not comprehensive in real-world scenarios, while average-based metrics (aADE and aFDE) can be more plausible and high level. To this end, we compared the proposed Multiclass-SGCN with Semantic-STGCNN using aADE and aADE (\tablename~\ref{tab:sotacomparison}), demonstrating a significant improvement of more than minus 10 for both metrics.

\begin{table}[t]
\caption{Performance comparison with the state-of-the-arts.}
\label{tab:allmodels}
\centering
\resizebox{0.60\columnwidth}{!}{
\begin{tabular}{lll}
\toprule

Model                                                           & mADE & mFDE \\
\midrule
Linear                                                          & 37.11& 63.51    \\
SF \cite{Yamaguchi2011bm}                                       & 36.48& 58.14    \\
Social-LSTM \cite{Alexandre2016lstm}                            & 31.19& 56.97    \\
Social-GAN \cite{gupta2018social}                               & 27.25& 41.44    \\
CAR-Net \cite{sadeghian2017carnet}                              & 25.72& 51.80    \\
DESIRE \cite{lee2017desire}                                     & 19.25& 34.05    \\
Social-STGCNN \cite{Mohamed2020socialstgcnn}                    & 26.46& 42.71    \\
Semantics-STGCNN \cite{rainbow2021semanticsstgcnn}              & 18.12& 29.70    \\
\midrule
Multiclass-SGCN (ours)                        & \textbf{14.36}& \textbf{25.99}    \\
\bottomrule
\end{tabular}
}
\end{table}

\begin{table}[t]
\caption{Performance comparison with Semantics-STGCNN.}
\label{tab:sotacomparison}
\centering
\resizebox{0.8\columnwidth}{!}{
\begin{tabular}{lllll}
\toprule
Model                                                           &mADE &mFDE &aADE &aFDE \\
\midrule
Semantics-STGCNN \cite{rainbow2021semanticsstgcnn}              &18.12 &29.70 &33.14 &61.14     \\
Multiclass-SGCN (ours)                        & \textbf{14.36}&\textbf{25.99} &\textbf{22.87} &\textbf{45.30}  \\
\bottomrule
\end{tabular}
}
\vspace{-5mm}
\end{table}

\begin{table}[t]
\caption{Ablation study results.}
\label{tab:ablation}
\centering
\resizebox{0.8\columnwidth}{!}{
\begin{tabular}{lcccc}
\toprule
Model                                       &mADE &mFDE &aADE &aFDE         \\
\midrule
Multiclass-SGCN w/o SE                                      &14.77 &\textbf{25.44} &24.74 &48.42    \\
Multiclass-SGCN w/o CL                                   &15.32	&26.39	 &26.29   &50.30   \\

Multiclass-SGCN w/o AIM                                     &22.05 &29.53 &40.33 &76.99    \\

Multiclass-SGCN (ours)                      &\textbf{14.36} &25.99 &\textbf{22.87} &\textbf{45.30} \\
\bottomrule
\end{tabular}
}
\end{table}

To further validate the contribution of class labels (CL), separate embedding (SE) of the VLG, and adaptive interaction mask (AIM), we conducted three ablation experiments by evaluating three variants of the proposed method: i) Mutliclass-SGCN w/o SE denotes that the embedding of the input graph was computed from the whole feature matrix, instead of separately for velocity and labels (Section \ref{sec:embedding}); ii) Mutliclass-SGCN w/o AIM denotes that a manually set interaction threshold ($\xi = 0.5$) was used for all agents to measure the existence of their neighbours, as in SGCN \cite{shi2021sgcn}, instead of our proposed adaptive interaction mask (Section \ref{sec:aim}); iii) Mutliclass-SGCN w/o CL denotes that the embedding of the input graph was computed only for velocity, instead of both velocity and class labels. Results in \tablename~\ref{tab:ablation} show that the proposed use of class labels and of the SE and AIM modules is important for boosting the performance of the model, especially AIM, which led to a 43.3\% reduction in aADE and a 41.2\% reduction in aFDE, indicating the importance of adaptively modelling the interaction patterns of each agent,
because agents of different classes may have different attention preferences.

\subsection{Qualitative Results}

\begin{figure}[t]
\centering
\begin{tabular}{c@{\hspace{0pt}}c@{\hspace{0pt}}c}
\includegraphics[width=0.32\columnwidth]{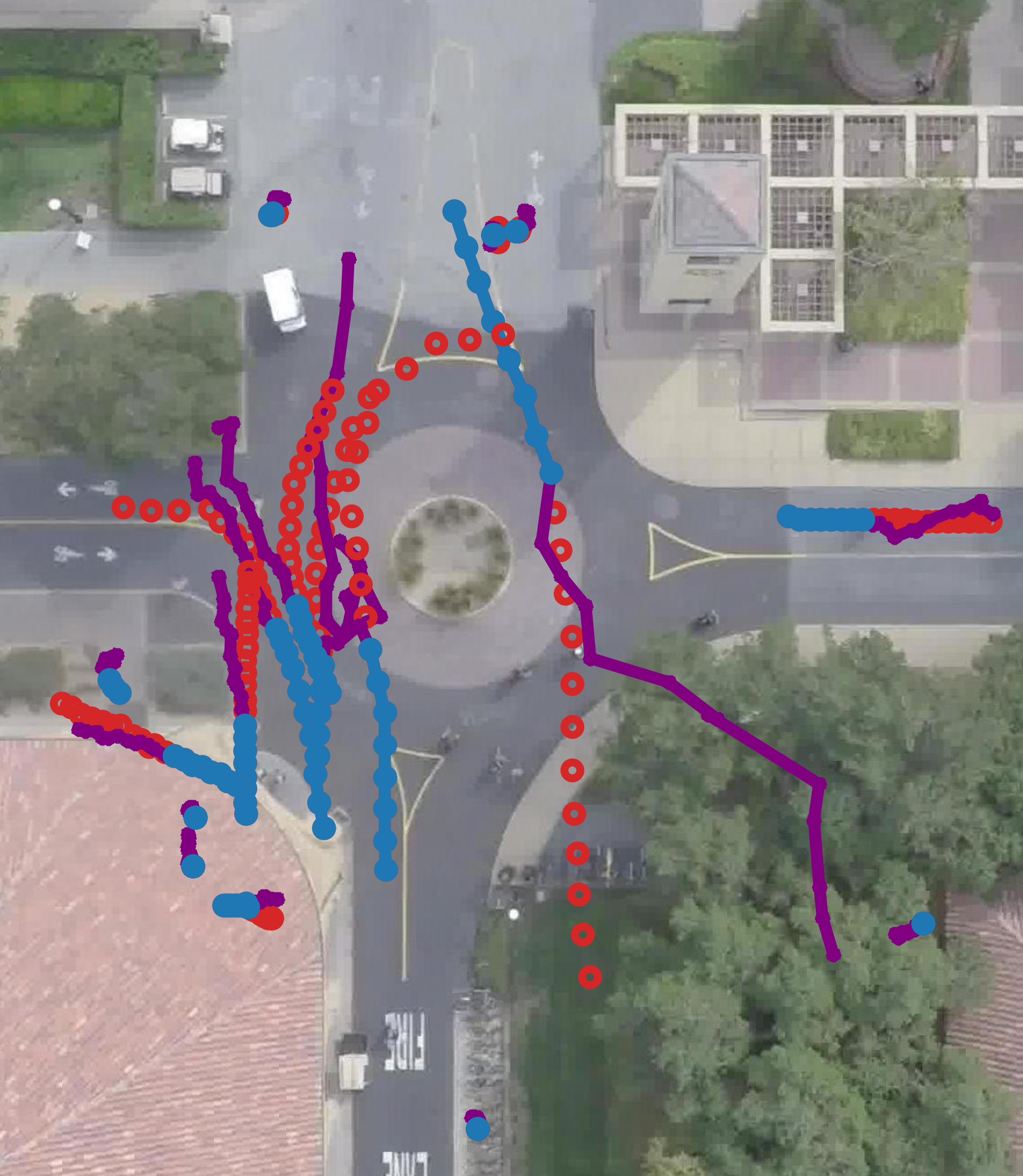}     &  \includegraphics[width=0.32\columnwidth]{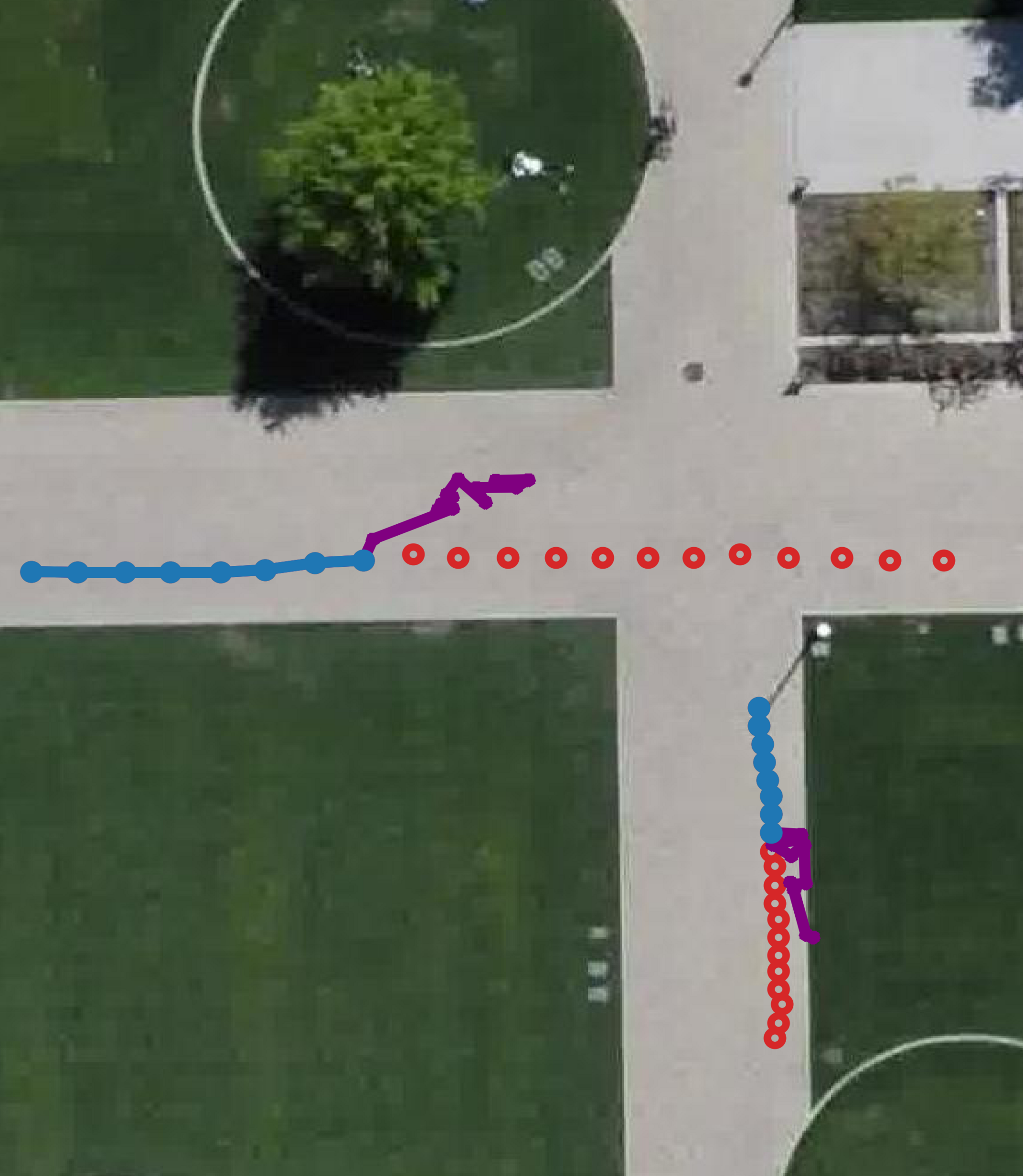} &  \includegraphics[width=0.32\columnwidth]{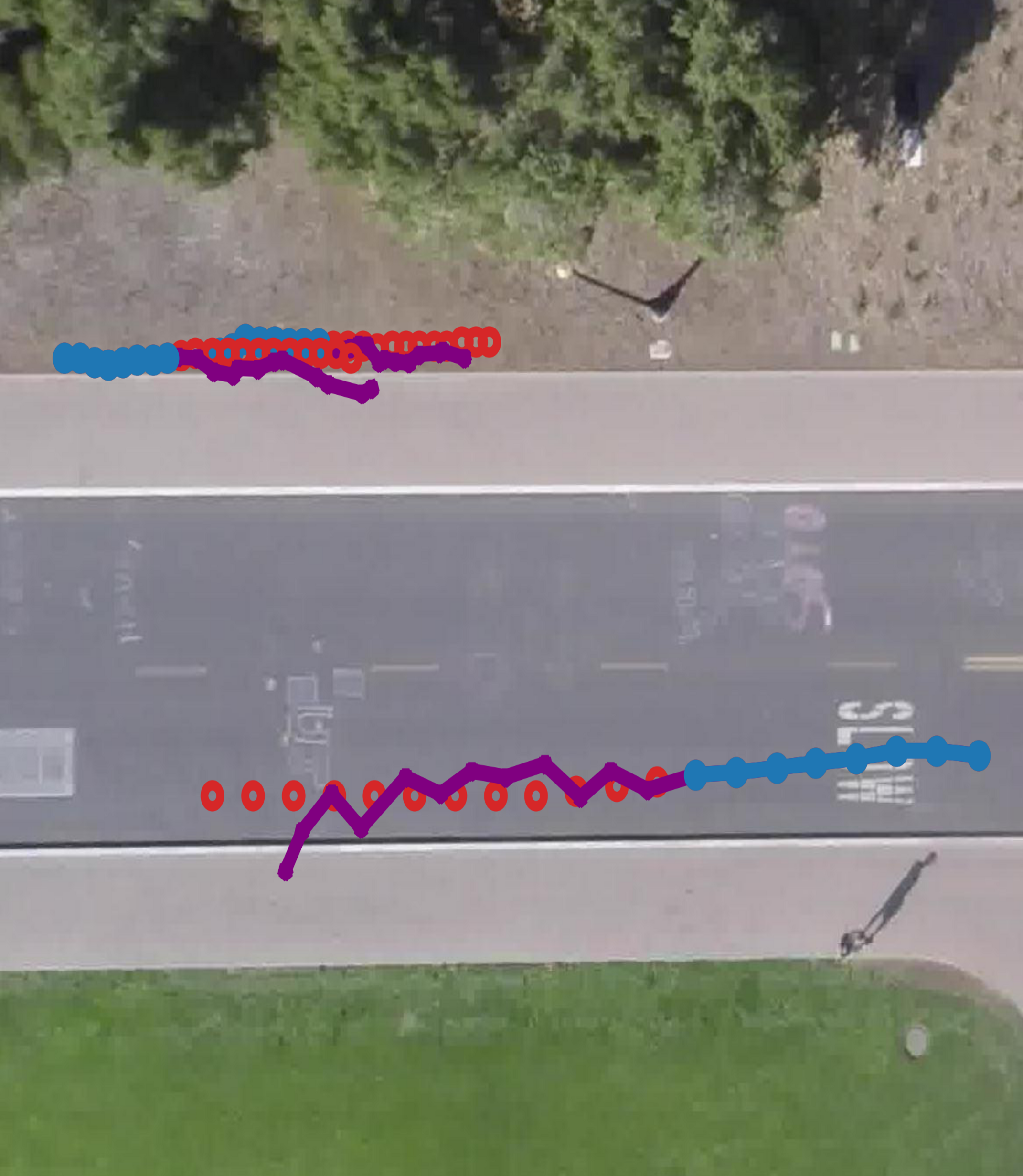} \\
\multicolumn{3}{c}{(a) Semantics-STGCNN \cite{rainbow2021semanticsstgcnn}} \\
\includegraphics[width=0.32\columnwidth]{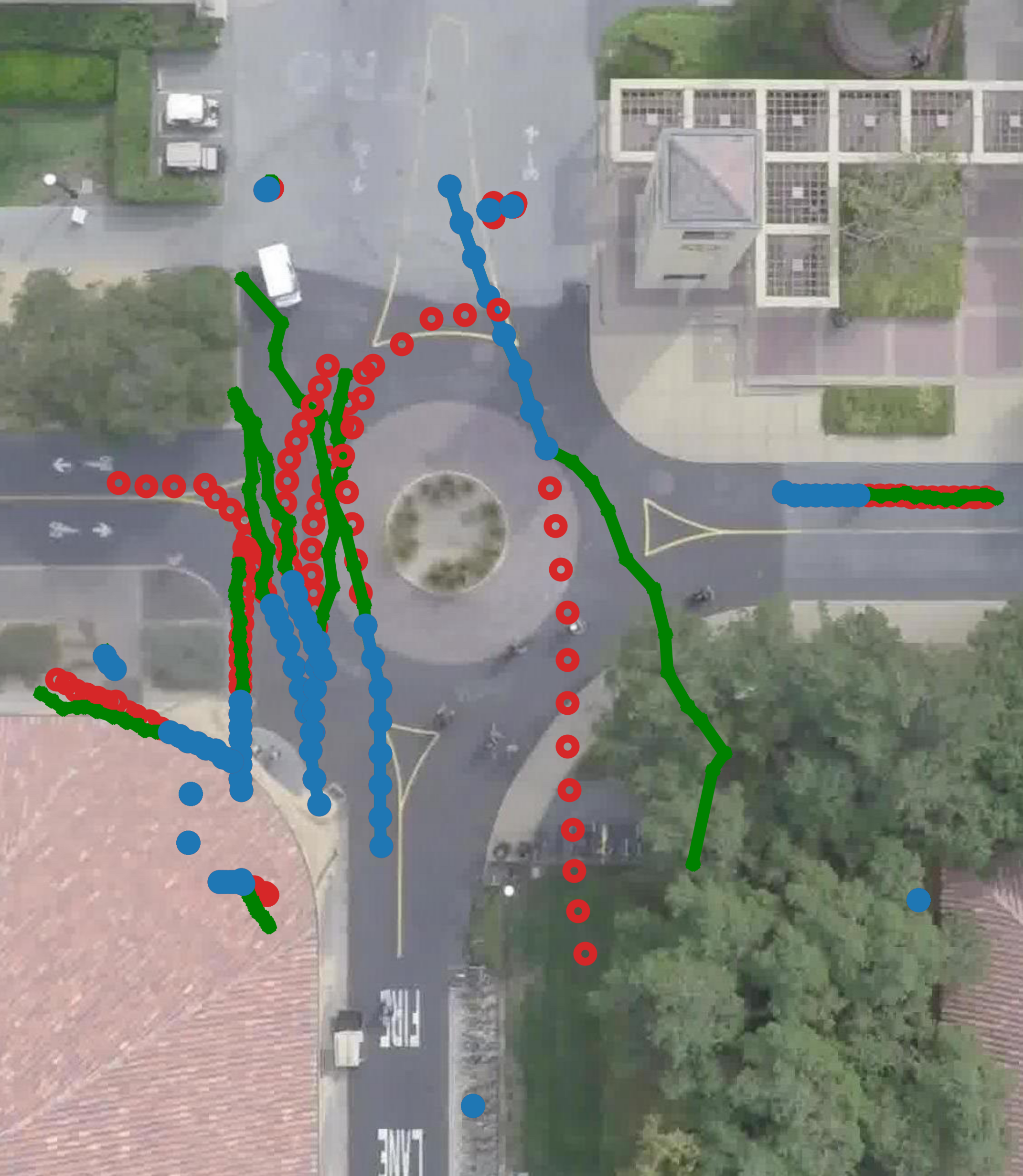}     &  \includegraphics[width=0.32\columnwidth]{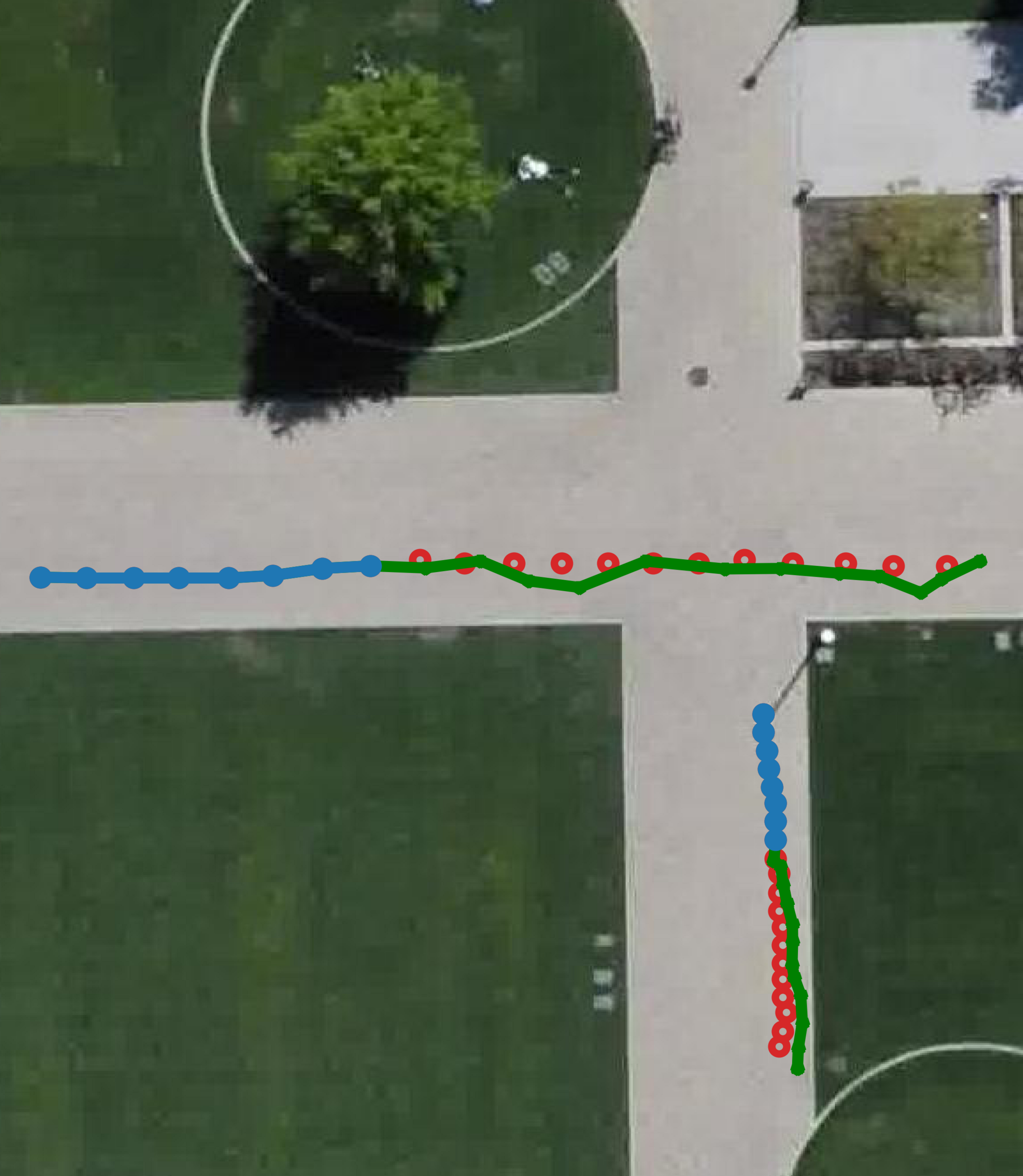} &  \includegraphics[width=0.32\columnwidth]{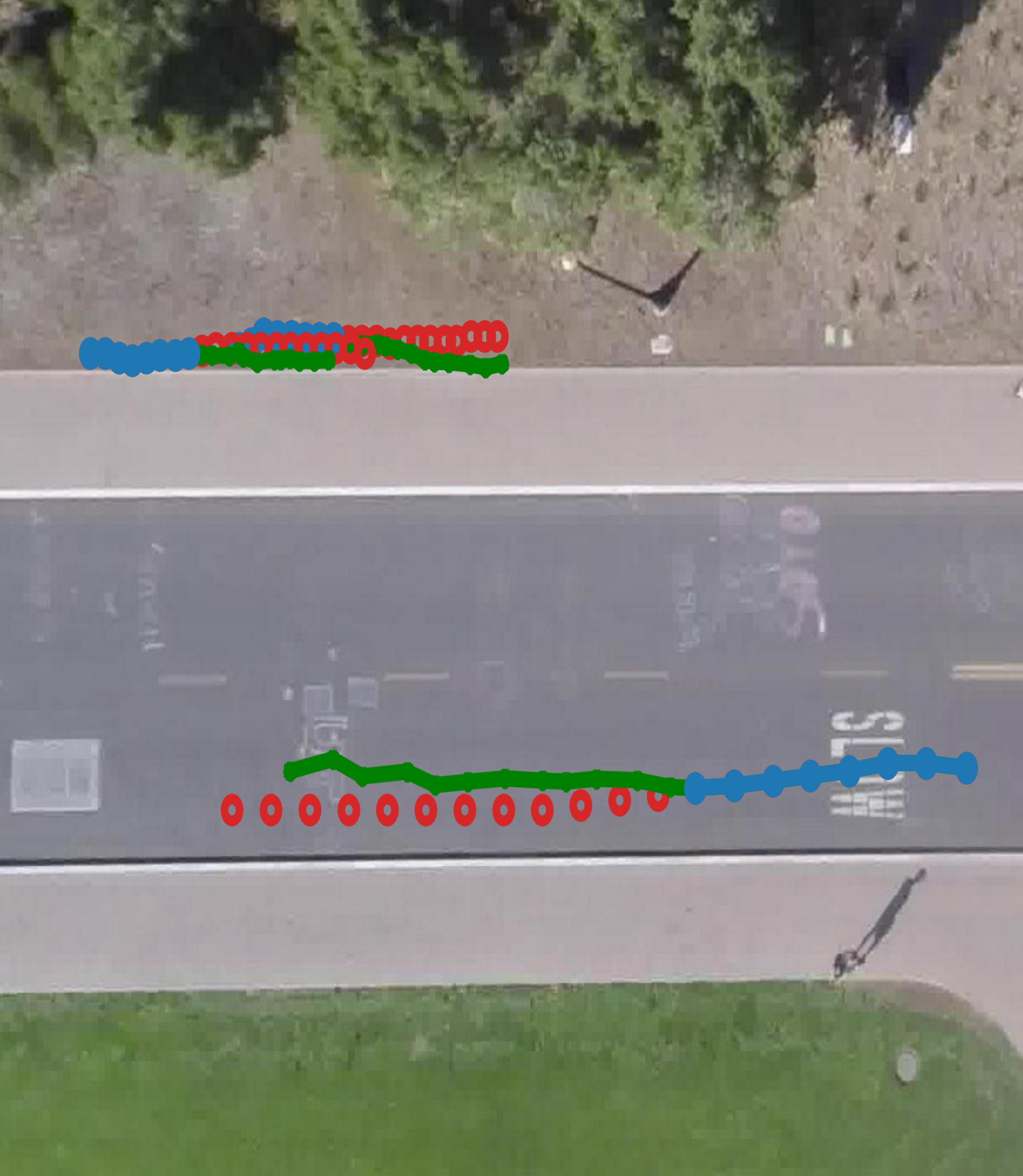} \\
\multicolumn{3}{c}{(b) Muticlass-SGCN (ours)} \\
\end{tabular}
\caption{\cite{rainbow2021semanticsstgcnn} vs. ours for single frames. Blue filled circles are observed trajectories, red hollow circles are ground-truth, purple lines in (a) are predicted results by \cite{rainbow2021semanticsstgcnn}, green lines in (b) are predicted results by the proposed Multiclass-SGCN.}
\label{fig:res}
\vspace{-3mm}
\end{figure}

Predicted trajectories by the proposed Multiclass-SGCN and Semantics-STGCNN~\cite{rainbow2021semanticsstgcnn} for one frame from three scenarios are shown in \figurename~\ref{fig:res}, demonstrating that our proposed model can make more realistic and consistent trajectory predictions.
Specifically, in the complex circular scenario (left-most images in \figurename~\ref{fig:res}), which contains too many agents, both methods failed to converge to the ground-truth, especially when agents are turning or moving at high speeds, but the prediction results of our Multiclass-SGCN exhibit less divergence and are better aligned with the ground-truth trajectories. Moreover, for some static agents, Semantics-STGCNN generates abnormal predictions, while our model does not. 
As for the middle images in \figurename~\ref{fig:res}, it is clear that Semantics-STGCNN totally diverges from the ground-truth, whereas our results match the ground-truth considerably.
Furthermore, for the right-most images in \figurename~\ref{fig:res}, both methods are close to the ground-truth, but Multiclass-SGCN presents more stable trajectories with lower amplitude oscillations.

To summarise, Semantics-STGCNN underperforms because the densely connected graph inherently introduces superfluous interactions that disrupt normal trajectories, and the lack of separate modelling of temporal dependencies results in unstable movements, even when no social interactions occur. In contrast, Multiclss-SGCN overcomes these issues by 
modelling both spatial interactions and temporal dependencies with velocity-label graph embedding and enhanced sparse graph learning modules, leading to better predictions.

\section{Conclusion}
\label{sec:conlusion}
In this paper, we introduced Multiclass-SGCN for multi-class trajectory predictions. To this end, we proposed the velocity-label graph that fuses velocity and label information in order to take into consideration different types of agents. We also designed a novel adaptive interaction mask to filter the high-level neighbours of each agent to maintain sparsity and enhance reliability. The experimental evaluation on the Stanford Drone Dataset demonstrated that our proposed method outperforms state-of-the-art approaches for all metrics considered. In the future, we intend to examine the integration of environmental features \cite{mangalam2021ynet} into our Multiclass-SGCN model to further improve prediction accuracy.

\bibliographystyle{IEEEbib}
\bibliography{main}

\begin{thebibliography}{10}

\bibitem{liu2021trajectorysurvey}
J.~Liu, X.~Mao, Y.~Fang, D.~Zhu, and M.~Meng,
\newblock ``A survey on deep-learning approaches for vehicle trajectory
  prediction in autonomous driving,''
\newblock {\em CoRR}, vol. abs/2110.10436, 2021.

\bibitem{zhou2015dynamic}
B.~Zhou, X.~Tang, and X.~Wang,
\newblock ``Learning collective crowd behaviors with dynamic
  pedestrian-agents,''
\newblock {\em International Journal of Computer Vision}, vol. 111, no. 1, pp.
  50--68, 2015.

\bibitem{Alexandre2016lstm}
A.~Alahi, K.~Goel, V.~Ramanathan, A.~Robicquet, L.~Fei-Fei, and S.~Savarese,
\newblock ``{Social LSTM}: Human trajectory prediction in crowded spaces,''
\newblock in {\em IEEE CVPR}, 2016, pp. 961--971.

\bibitem{Mohamed2020socialstgcnn}
M.~Abduallah, Q.~Kun, E.~Mohamed, and C.~Christian,
\newblock ``Social-stgcnn: A social spatio-temporal graph convolutional neural
  network for human trajectory prediction,''
\newblock in {\em IEEE/CVF CVPR}, 2020, pp. 14412--14420.

\bibitem{bai2018tcn}
S.~Bai, J.~Z. Kolter, and V.~Koltun,
\newblock ``An empirical evaluation of generic convolutional and recurrent
  networks for sequence modeling,''
\newblock {\em CoRR}, vol. abs/1803.01271, 2018.

\bibitem{YuMa2020Spatio}
C.~Yu, X.~Ma, J.~Ren, H.~Zhao, and S.~Yi,
\newblock ``Spatio-temporal graph transformer networks for pedestrian
  trajectory prediction,''
\newblock in {\em ECCV}, August 2020.

\bibitem{giuliari2020Transformer}
I.~Giuliari, F.and~Hasan, M.~Cristani, and F.~Galasso,
\newblock ``Transformer networks for trajectory forecasting,''
\newblock in {\em ICPR}, 2021, pp. 10335--10342.

\bibitem{Vaswani2017Transformer}
A.~Vaswani, N.~Shazeer, N.~Parmar, J.~Uszkoreit, L.~Jones, A.N. Gomez,
  L.~Kaiser, and I.~Polosukhin,
\newblock ``Attention is all you need,''
\newblock in {\em NIPS}, 2017.

\bibitem{shi2021sgcn}
L.~Shi, L.~Wang, C.~Long, S.~Zhou, M.~Zhou, Z.~Niu, and G.~Hua,
\newblock ``{SGCN}: Sparse graph convolution network for pedestrian trajectory
  prediction,''
\newblock in {\em IEEE/CVF CVPR}, 2021, pp. 8990--8999.

\bibitem{rainbow2021semanticsstgcnn}
B.~A. Rainbow, Q.~Men, and H.~P.~H. Shum,
\newblock ``{Semantics-STGCNN}: A semantics-guided spatial-temporal graph
  convolutional network for multi-class trajectory prediction,''
\newblock in {\em IEEE SMC}, 2021, pp. 2959--2966.

\bibitem{men21pytorch}
Qianhui Men and Hubert P.~H. Shum,
\newblock ``Pytorch-based implementation of label-aware graph representation
  for multi-class trajectory prediction,''
\newblock {\em Software Impacts}, vol. 11, pp. 100201, 2021.

\bibitem{zhang2019semanticskeleton}
P.~Zhang, C.~Lan, W.~Zeng, J.~Xing, J.~Xue, and N.~Zheng,
\newblock ``Semantics-guided neural networks for efficient skeleton-based human
  action recognition,''
\newblock in {\em IEEE/CVF CVPR}, 2020, pp. 1109--1118.

\bibitem{kipf2016gcn}
T.~N. Kipf and M.~Welling,
\newblock ``Semi-supervised classification with graph convolutional networks,''
\newblock in {\em ICLR}, 2017.

\bibitem{Robicquet2016SDD}
A.~Robicquet, A.~Sadeghian, A.~Alahi, and S.~Savarese,
\newblock ``Learning social etiquette: Human trajectory understanding in
  crowded scenes,''
\newblock in {\em ECCV}, 2016, pp. 549--565.

\bibitem{Szegedy2015AsyConv}
C.~Szegedy, V.~Vanhoucke, S.~Ioffe, J.~Shlens, and Z.~Wojna,
\newblock ``Rethinking the inception architecture for computer vision,''
\newblock in {\em IEEE CVPR}, 2016, pp. 2818--2826.

\bibitem{gupta2018social}
A.~Gupta, J.~Johnson, L.~Fei-Fei, S.~Savarese, and A.~Alahi,
\newblock ``Social gan: Socially acceptable trajectories with generative
  adversarial networks,''
\newblock in {\em IEEE/CVF CVPR}, 2018, pp. 2255--2264.

\bibitem{kingma2014method}
D.~P. Kingma and J.~Ba,
\newblock ``Adam: A method for stochastic optimization,''
\newblock in {\em ICLR}, 2015.

\bibitem{Yamaguchi2011bm}
K.~Yamaguchi, A.~C. Berg, L.~E. Ortiz, and T.~L. Berg,
\newblock ``Who are you with and where are you going?,''
\newblock in {\em CVPR}, 2011, pp. 1345--1352.

\bibitem{sadeghian2017carnet}
A.~Sadeghian, F.~Legros, M.~Voisin, R.~Vesel, A.~Alahi, and S.~Savarese,
\newblock ``Car-net: Clairvoyant attentive recurrent network,''
\newblock in {\em ECCV}, 2018, pp. 162--180.

\bibitem{lee2017desire}
N.~Lee, W.~Choi, P.~Vernaza, C.~B. Choy, P.~H.~S. Torr, and M.~Chandraker,
\newblock ``{DESIRE}: {D}istant future prediction in dynamic scenes with
  interacting agents,''
\newblock in {\em IEEE CVPR}, 2017, pp. 2165--2174.

\bibitem{mangalam2021ynet}
K.~Mangalam, Y.~An, H.~Girase, and J.~Malik,
\newblock ``From goals, waypoints \& paths to long term human trajectory
  forecasting,''
\newblock in {\em IEEE/CVF ICCV}, October 2021, pp. 15233--15242.

\end{thebibliography}

\end{document}